\documentclass[10pt,journal,compsoc]{IEEEtran}
\usepackage{times}
\usepackage{epsfig}
\usepackage{graphicx}
\usepackage{amsmath}
\usepackage{amssymb}
\usepackage{subfigure}
\usepackage{epstopdf}
\usepackage{chngpage}
\usepackage{array}
\usepackage{makecell}
\usepackage{multirow}
\usepackage[noend]{algpseudocode}
\usepackage{algorithmicx}
\usepackage{algorithm}
\usepackage{url}
\usepackage{booktabs}
\usepackage{color}
\usepackage{caption}
\usepackage{amssymb}
\usepackage{ragged2e}

\ifCLASSOPTIONcompsoc
  \usepackage[nocompress]{cite}
\else
  \usepackage{cite}
\fi
\ifCLASSINFOpdf
\else
\fi
\begin{document}
%

\title {Edge-Semantic Learning Strategy for Layout Estimation in Indoor Environment}

\author{Weidong~Zhang$^*$,~\IEEEmembership{Student~Member,~IEEE,}
        Wei~Zhang$^*$,~\IEEEmembership{Member,~IEEE,}
        and~Jason~Gu,~\IEEEmembership{Senior~Member,~IEEE}
\IEEEcompsocitemizethanks{\IEEEcompsocthanksitem W.D. Zhang, W. Zhang, and J. Gu are with the School of Control Science and Engineering, Shandong University, Jinan, Shandong, 250061 China.\protect\\
E-mail: davidzhangsdu@gmail.com \protect\\
$^*$ The first two authors contributed equally.
}}

\IEEEtitleabstractindextext{%
\begin{abstract}
Visual cognition of the indoor environment can benefit from the spatial layout estimation, which is to represent an indoor scene with a 2D box on a monocular image. In this paper, we propose to fully exploit the edge and semantic information of a room image for layout estimation. More specifically, we present an encoder-decoder network with shared encoder and two separate decoders, which are composed of multiple deconvolution (transposed convolution) layers, to jointly learn the edge maps and semantic labels of a room image. We combine these two network predictions in a scoring function to evaluate the quality of the layouts, which are generated by ray sampling and from a predefined layout pool. Guided by the scoring function, we apply a novel refinement strategy to further optimize the layout hypotheses. Experimental results show that the proposed network can yield accurate estimates of edge maps and semantic labels. By fully utilizing the two different types of labels, the proposed method achieves state-of-the-art layout estimation performance on benchmark datasets.
\end{abstract}

}

\maketitle

\IEEEdisplaynontitleabstractindextext

%
\IEEEpeerreviewmaketitle

\IEEEraisesectionheading{\section{Introduction}\label{sec:introduction}}

%
%
%
%

\IEEEPARstart{T}{he} indoor scene is a basic environment for vision and robotics. As an important property of indoor scenes, the spatial layout carries the three-dimensional (3D) information, and can provide a better understanding of indoor scenes. The spatial layout can serve as geometric constraints for various tasks such as indoor modeling~\cite{karsch2011rendering,xiao2014reconstructing,martin20143d,liu2015rent3d,Camplani2013Depth}, indoor navigation~\cite{joseph2013semantic, Li2017A}, and visual cognition~\cite{hedau2010thinking,Qiao2016Biologically}. Specifically, the problem of spatial layout estimation is to find a 2D box to represent an indoor scene. For a monocular indoor image, the problem is to locate the wall-floor, wall-wall, and wall-ceiling boundaries. The layout can also be represented by pixel-wise segmentation, with different labels representing different semantic surfaces (i.e., the ceiling, floor, front wall, left wall, and right wall).

\begin{figure}
\begin{center}
\includegraphics[width = 1\columnwidth]{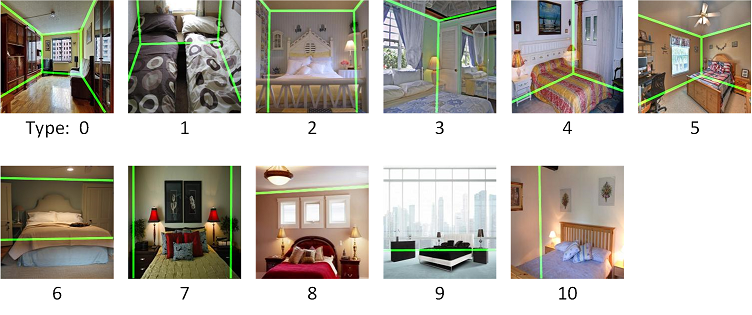}
\end{center}
\caption{Definition of the 11 layout types in the LSUN Challenge~\cite{yinda2016lsun}.}
\label{fig:type}
\end{figure}

The ability to estimate the spatial layout from a single image remains a challenging task. First, it is difficult to locate the desired boundaries of the wall and floor as they are always mixed with the edges of cluttered indoor objects, and are often occluded severely. There is a similar problem with respect to estimating the semantic labels. It is challenging to describe all variations of layouts in a parametric way, some pioneering work~\cite{lee2017roomnet} was presented recently though. This is because the indoor scenes are captured from various viewpoints, and thus there may be very different types of layouts, as shown in Fig.~\ref{fig:type}.


\begin{figure*}
\begin{center}
\includegraphics[width = 1.0\textwidth]{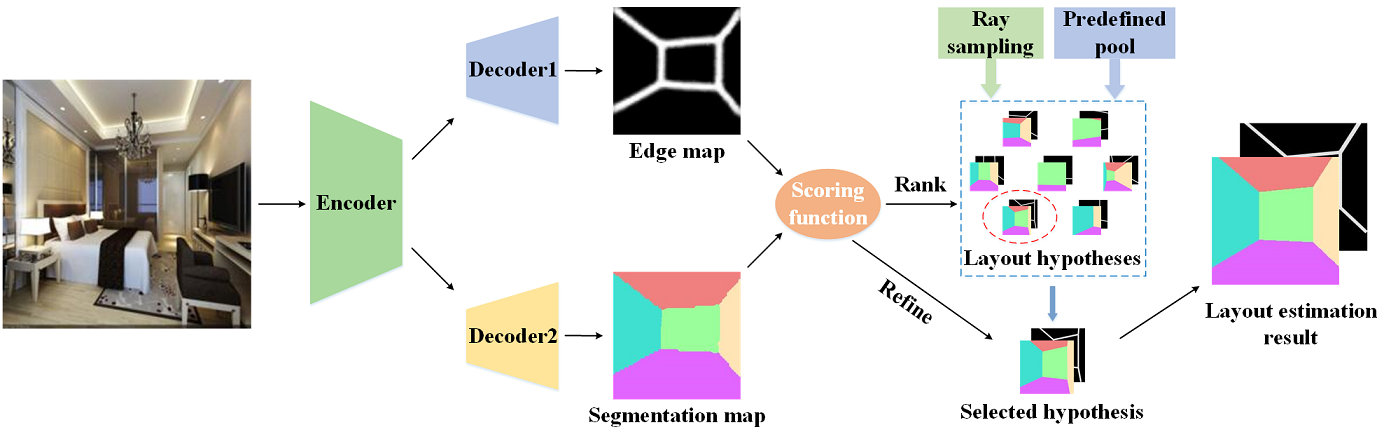}
\end{center}
\caption{An overview of the layout-estimation pipeline of our method. The proposed network has a shared encoder and two separate decoders to predict the edge map and segmentation map. These two predictions are combined into a scoring function to select and refine the layout hypotheses, which are generated by ray sampling and from a predefined layout pool.}
\label{fig:overview}
\end{figure*}

The problem would be tractable if heat maps representing the room edges or semantic surfaces could be precisely estimated. For clarity, the names of such heat maps are defined as follows: An \textbf{edge map} is a heat map that represents the boundaries of the ceiling, walls, and the floor. The \textbf{semantic labels} are five belief maps, each of which represents the region of a semantic surface of the room. In turn, the five semantic surfaces comprise the ceiling, floor, front wall, left wall, and right wall. The semantic labels can be converted to a single \textbf{segmentation map}, which is a labeling map that represents the semantic surfaces with different labels.

Existing work has relied only on either edge maps~\cite{mallya2015learning,ren2016coarse} or semantic labels~\cite{dasgupta2016delay} for layout estimation, which are all learned by fully convolutional neural networks (FCNs)~\cite{long2015fully1}. Although the semantic information was included for multi-task learning in \cite{mallya2015learning} and \cite{ren2016coarse}, they were only used to aid the training of the edge maps, and did not add any benefit after training the networks. Therefore, the study about the combination of the edge maps and semantic labels for layout estimation is rarely studied in the literature. Moreover, it is found that the estimated label maps produced by the FCNs are of low quality, and are unreliable. 

To our knowledge, this work is the first attempt to ``truly'' use the edge and semantic information for layout estimation, i.e., the two estimates are jointly learned not only to enhance the training, but also to compensate each other for layout inference. Fig.~\ref{fig:overview} gives an overview of the layout-estimation pipeline. Unlike existing FCN-based methods, we designed an encoder-decoder network to jointly learn the edge map and semantic labels from a single room image. The proposed network has two separate decoders that are composed of multiple deconvolution (transposed convolution) layers. Compared to FCNs, the parallel and hierarchical structure of deconvolution layers enables the model to generate high-quality edge maps and semantic labels simultaneously in a coarse-to-fine manner. Extensive results on benchmark datasets demonstrate the benefits of the proposed joint learning strategy on both training and layout prediction.

To exploit the similarity of interior spatial organization, we proposed a novel strategy for generating layout proposals by searching in a predefined pool. This is also new in the literature, as most existing methods rely on vanishing point detection, which is sensitive to noise and hard to estimate in complex cluttered rooms. The use of predefined layouts collected from existing instances significantly enhances the robustness. Finally, we performed an optimization step on the layout hypotheses to yield the final layout estimates.  

The main contributions of this paper are summarized as follows:
1) We present a framework to combine the edge maps and semantic labels for layout estimation. It is found that joint learning the edge maps and semantic labels mutually benefit each other. More importantly, the estimated edge maps and semantic labels proved to compensate each other for layout inference.
2) We propose a novel strategy for layout proposal generation, and a pixel-level optimization method for layout refinement. The results show that the proposed method outperformed state-of-the-art methods with significant accuracy gains.


\section{Related work}


\begin{figure*}
\begin{center}
\includegraphics[width = 1.0\textwidth]{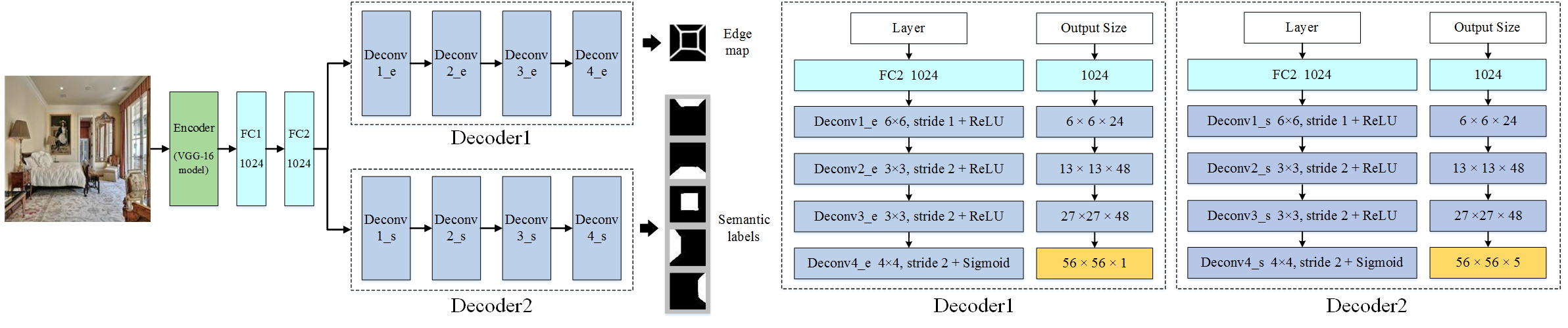}
\end{center}
\caption{Architecture of the proposed network. We employed the VGG-16 model (from conv1$\_$1 layer to conv5$\_$3 layer)~\cite{simonyan2014very} as the encoder.}
\label{fig:net}
\end{figure*}

The task of spatial layout estimation was first introduced by Hedau et al.~\cite{hedau2009recovering}. First, a series of layout hypotheses were generated by uniformly sampling rays from the vanishing points. Then, each layout hypothesis was assigned a score by a learned structured regressor, and the layout with the highest score was selected as the final result. Schwing et al.~\cite{schwing2012efficient,schwing2012efficient1} employed dense sampling and used integral geometry decomposition for efficient structure prediction. Lee et al.~\cite{lee2009geometric} evaluated the layout hypotheses based on the consistency of the layouts to the orientation maps, which were computed from line segments and represent the local belief of region orientations. Wang et al.~\cite{wang2013discriminative} considered the indoor clutters and modeled the room faces and clutter layouts jointly with latent variables. Pero et al.~\cite{del2013understanding,pero2012bayesian} used a generative model that aggregates the 3D geometry and indoor objects. MCMC sampling was performed to search the model parameters.

Recently, edge maps or semantic labels learned from FCNs~\cite{long2015fully1} have become popular for this task, and their use has significantly enhanced the layout estimation performance. Mallya and Lazebnik~\cite{mallya2015learning} used the FCN to predict the edge maps. An adaptive vanishing line sampling method based on the learned edge maps was proposed to generate the layout hypotheses. Then, the hypotheses were ranked by the learned edge maps together with the line membership (LM)~\cite{hedau2009recovering} and geometric context (GC) features~\cite{hoiem2007recovering}. Dasgupta et al.~\cite{dasgupta2016delay} used the FCN to predict the semantic labels. The initial layout hypothesis was generated by logistic regression, and was further optimized by the four vanishing lines and one vanishing point. Ren et al.~\cite{ren2016coarse} used the learned edge maps as a reference for generating layout hypotheses based on the vanishing lines, undetected lines, and occluded lines. In addition, Lee et al.~\cite{lee2017roomnet} adopted a direct formulation of this problem by predicting the locations of the room layout keypoints with an end-to-end network. Zhao et al.~\cite{zhao2017physics} transferred the semantic segmentation to edge estimates and proposed the physics inspired optimization inference scheme.

\begin{figure}
\begin{center}
\subfigure{
\includegraphics[width = .35\columnwidth]{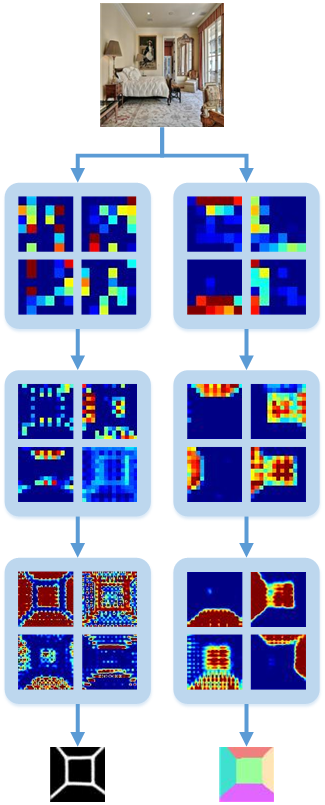}
}
\subfigure{
\includegraphics[width = .35\columnwidth]{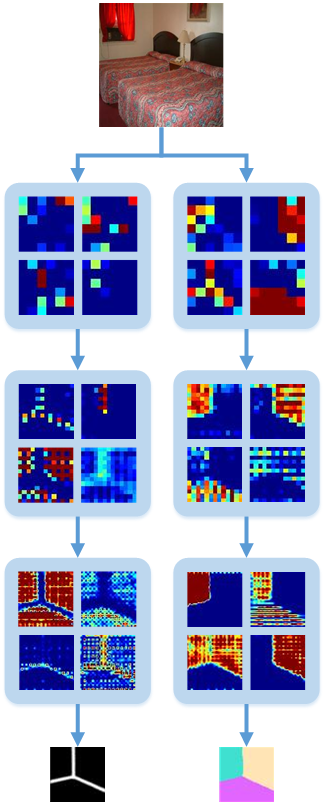}
}
\end{center}
\caption{Activation visualization of the proposed network. The activations from top to bottom are the outputs from the first, second, and third deconvolution layers, respectively. Four feature maps are randomly selected for each layer. The original sizes of the intermediate feature maps are 6$\times$6, 13$\times$13, and 27$\times$27.}
\label{fig:visual}
\end{figure}

In~\cite{mallya2015learning} and~\cite{ren2016coarse}, the edge maps were jointly learned with semantic labels. However, the essence and purpose differ much from ours. Although the semantic information was included in~\cite{mallya2015learning} and~\cite{ren2016coarse}, they were only used to aid the training of the edge maps, and did not add any benefit after training the networks. The reason might be the lack of deep decoders as in FCN~\cite{long2015fully1, noh2015learning}. That is, given the edge output, not much additional information is provided for layout estimates by the semantic output, as stated in~\cite{mallya2015learning} and~\cite{ren2016coarse}. So, the semantic outputs are abandoned in the layout estimation in both~\cite{mallya2015learning} and~\cite{ren2016coarse}. By contrast, what we emphasize here is the first attempt to “truly” use the two information (edge and semantic information) for layout estimation, i.e., the two outputs are jointly learned not only to enhance the training, but also to compensate each other for layout estimation. As shown in Fig.~\ref{fig:net}, the two types of outputs are estimated with shared encoder and two separate decoders composed of hierarchical deconvolution layers. The benefits are threefold: 1) The high-level features benefitted from the two-stream joint learning structure might be learned from the shared encoder; 2) The layout estimates can be generated gradually in the decoders in a coarse-to-fine manner (see Fig.~\ref{fig:visual}). Consequently, the edge and semantic estimates are more accurate than those of existing work (see Fig.~\ref{fig:fcncmp} and Fig.~\ref{fig:seg_comparison}); 3) As the two decoders are separate and have deep structure, the corresponding two types of outputs are therefore less correlated and could compensate each other for layout estimation. The results in Table~\ref{table:train} show that joint training and combine the two outputs provide a considerable boost in layout accuracy. The layout estimates are good even one type of outputs are poor as shown Fig.~\ref{fig:bad} (a) and (b). To our knowledge, this work is the first one that makes edge and semantic information applicable in layout estimation, while the semantic information was abandoned in traditional work\cite{mallya2015learning, ren2016coarse} as aforementioned.

\begin{figure*}[t]
\begin{center}
\subfigure[]{
\includegraphics[scale = 0.32]{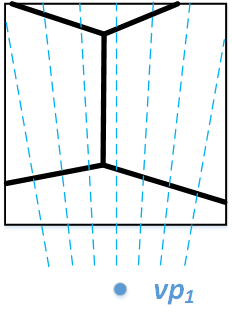}
}
\subfigure[]{
\includegraphics[scale = 0.32]{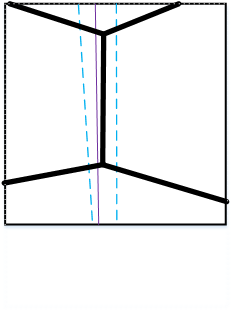}
}
\subfigure[]{
\includegraphics[scale = 0.32]{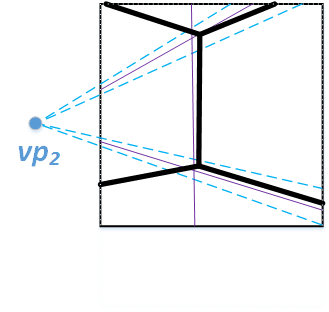}
}
\subfigure[]{
\includegraphics[scale = 0.32]{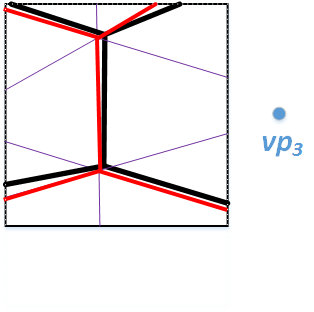}
}
\subfigure[]{
\includegraphics[scale = 0.32]{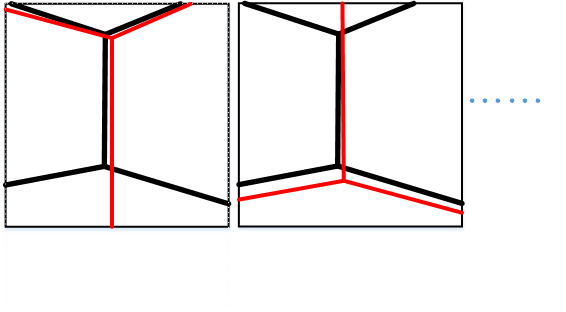}
}
\end{center}
\caption{Illustration of the layout generation by ray sampling. (a) Uniformly spaced sectors generated from the vertical vanishing point $vp_1$. (b) The sector with the local maximum edge strength is selected and a ray (outlined by purple) is sampled in the selected sector. (c) The selected sectors and sampled rays of the farther horizontal vanishing point $vp_2$. (d) A layout candidate (outlined by red) obtained by generating rays from $vp_3$ to go through the intersection points of the previously sampled rays of $vp_1$ and $vp_2$. (e) Layouts generated with different types by using a subset of the selected sectors.}
\label{fig:samp}
\end{figure*}

\section{Estimating room edges and pixel semantic labels}
\label{sec:net}

The most decisive factor for the final performance is estimating the edge maps and semantic labels. In this work, an encoder-decoder network is trained for layout estimation. The architecture of the network is shown in Fig.~\ref{fig:net}. The encoder is the same with the VGG-16 model (from conv1$\_$1 layer to conv5$\_$3 layer)~\cite{simonyan2014very}. Two fully connected layers of size 1024 are followed subsequently. Then, the network is divided into two branches of decoders for different outputs. Each decoder has four successive deconvolution layers and ends with sigmoid activation. ReLU~\cite{krizhevsky2012imagenet} activation is employed between the fully connected layers and deconvolution layers.


During training, the input images are resized to $224\times224$ and fed to the network. For both branches, the output size is $56\times56$. The first branch is trained to predict the edge maps (as shown in Fig.~\ref{fig:net}). An edge map for training is first generated, and has the same size as the input image (for convenience of cropping), with six-pixel-wide lines to represent the edges. Besides, Gaussian blur (e.g., with $\sigma = 6$) is performed on the edge map to smoothen the boundaries of edges and non-edge regions. We employed both line thickening and Gaussian blur to enable the training loss to decrease easily and gradually as the predicted results approach the desired blurry regions. Then, we resized the edge map to $56\times56$. The output of the second branch consists of five heat maps. Each channel is trained to predict one specific semantic label. To determine the labelling, we used the definition in \cite{dasgupta2016delay} to avoid ambiguity of semantic labels: If only two walls are visible, they are labelled as a left wall and right wall; if only a single wall is visible, it is labelled as a front wall.

The total loss for training is the sum of the cross-entropy loss of both outputs. We used Caffe~\cite{jia2014caffe} to implement the network. The network is initialized with the weights of the VGG-16 model~\cite{simonyan2014very}, which is pretrained on the ILSVRC~\cite{deng2009imagenet} dataset. We set the learning rate to $10^{-4}$, and the momentum to $0.5$. 


To better understand the internal operations of the decoders, in Fig.~\ref{fig:visual}, we show the activations of the intermediate deconvolution layers. As can be seen, the coarse outlines of edges and semantic labels are discovered in the second deconvolution layer, and become clearer in the third layer. The visualization reveals that the multiple deconvolution layers can gradually refine the predicted heat maps in a coarse-to-fine manner. Consequently, the final outputs of the fourth deconvolution layer are visually accurate and clear.


\begin{table}[]
\centering
\caption{Comparison of the advantage and disadvantage of edge map and segmentation map for layout estimation.}
\label{table:summary}
\begin{tabular}{c|c|c|c|}
\cline{2-4}
                                                                                              & Edge map       & \begin{tabular}[c]{@{}c@{}}Segmentation\\ map\end{tabular} & \begin{tabular}[c]{@{}c@{}}Edge map +\\ Segmentation map\end{tabular} \\ \hline
\multicolumn{1}{|c|}{\begin{tabular}[c]{@{}c@{}}Discrete label\\ for each pixel\end{tabular}} &                & $\blacksquare$                                             & $\blacksquare$                                                        \\ \hline
\multicolumn{1}{|c|}{\begin{tabular}[c]{@{}c@{}}Score changes\\ smoothly\end{tabular}}        &                & $\blacksquare$                                             & $\blacksquare$                                                        \\ \hline
\multicolumn{1}{|c|}{\begin{tabular}[c]{@{}c@{}}No ambiguity\\ issue\end{tabular}}            & $\blacksquare$ &                                                            & $\blacksquare$                                                        \\ \hline
\end{tabular}
\end{table}

\section{Layout generation and refinement}
\label{sec:gene}

In this section, we aim to produce parameterized layout representation that can well reflect the estimated edges and semantic labels obtained from Section~\ref{sec:net}. We develop a comprehensive strategy for layout generation and refinement. First, we generated the layout hypotheses in two ways: (1) Adaptive ray sampling for enforcing the indoor geometric constraints; (2) A layout pool collected from existing samples to utilize the similarity of interior spatial organization. Finally, we refined the layout hypotheses via an optimization process to produce the layout estimate.

The size of the original images can vary greatly. For simplicity, we resized each of the input images as well as their edge map and semantic labels such that they are square with size $w\times w$. We then generated the parameterized layouts based on the square images. Once an optimal layout is obtained, its coordinate values are scaled to fit the original image size. In the implementation, $w$ is set to 224.  

For each image, the predicted edge map is denoted as $\textbf{E}$, and the semantic labels are denoted as $\textbf{m}^{i}$, $i \in \left [ 1,...,5 \right ]$. Both $\textbf{E}$ and $\textbf{m}^{i}$ are already resized to $w\times w$ via cubic interpolation and the image number is omitted for simplicity. We converted the five-channel semantic labels to a single segmentation map $\textbf{M}$:

\begin{equation}
\begin{aligned}
\label{eq:segment}
\textbf{M}(u,v) = \mathop{\arg}\mathop{\max}_{i} \textbf{m}^{i}(u,v),\forall u,v\in \left [ 1,...,w \right ].
\end{aligned}
\end{equation}

As shown in Fig.~\ref{fig:type}, we used the definition of the LSUN layout challenge~\cite{yinda2016lsun} to parameterize the layouts. In the definition, there are 11 types of room layouts that can cover most room images. The layout of each room is represented by the type to which it belongs and the coordinates of the corner points. Let $l = (t,p_{1},p_{2},...,p_{n})$ be a parameterized layout, where $t$ is the type of $l$ and $p_{1}$ to $p_{n}$ are the corner points of $l$. The role of each corner point in the sequence is determined by the type $t$. We defined $f$ and $g$ to be the functions that map $l$ to a homogeneous edge map $\textbf{E}_{l}$ and a segmentation map $\textbf{M}_{l}$, respectively. Similar to the training samples in Section \ref{sec:net}, $\textbf{E}_{l}$ is generated with six-pixel-wide lines and then smoothed by Gaussian blur with $\sigma=6$.

\begin{equation}
\begin{aligned}
\label{eq:f}
\textbf{E}_{l} = f(l),  \quad \textbf{M}_{l} = g(l).
\end{aligned}
\end{equation}

In the layout inference procedure, both edge and semantic information have favorable and unfavorable aspects as shown in Table~\ref{table:summary}. On one hand, the matching score between the semantic segmentation maps is about the dense pixel-wise accuracy, which changes smoothly and is beneficial to the layout inference procedure. In contrast, the detected edges are sparse. The score between two edge maps is derived from the Euclidean distance, which may not respond sensitively unless the corresponding edges are close. On the other hand, the edge map does not suffer from the ambiguity issue, while the segmentation map estimates often have disjoint components incurred by the labeling ambiguity. Therefore, using edge map or segmentation map alone is insufficient for layout estimation.

Therefore, in order to combine the two information together, the scoring function that reflects the consistency of $l$ with the estimated edge map $\textbf{E}$ and segmentation map $\textbf{M}$ is defined as follows:

\begin{equation}
\begin{aligned}
\label{eq:score}
S(\textbf{M}_{l},\textbf{E}_{l}|\textbf{M},\textbf{E}) = S_1(\textbf{M}_{l},\textbf{M}) + \lambda S_2(\textbf{E}_{l},\textbf{E}),
\end{aligned}
\end{equation}

\noindent where $S_1$ is the pixel-wise accuracy of the maximum bipartite matching~\cite{west2001introduction} of two segmentation maps. Specifically, for two segmentation maps, one from the candidate layout and one from the estimated segmentation map, the label consistency of any pairs of wall regions is first calculated as cost. Then the bipartite matching with the maximal cost is searched to denote the pixel-wise accuracy. Maximum bipartite matching has been adopted in the evaluation metrics of LSUN challenge~\cite{yinda2016lsun} to deal with the ambiguity among different corners and wall regions. $S_2$ denotes the negative Euclidean distance.


\begin{figure}
\begin{center}
\includegraphics[width = 0.8\columnwidth]{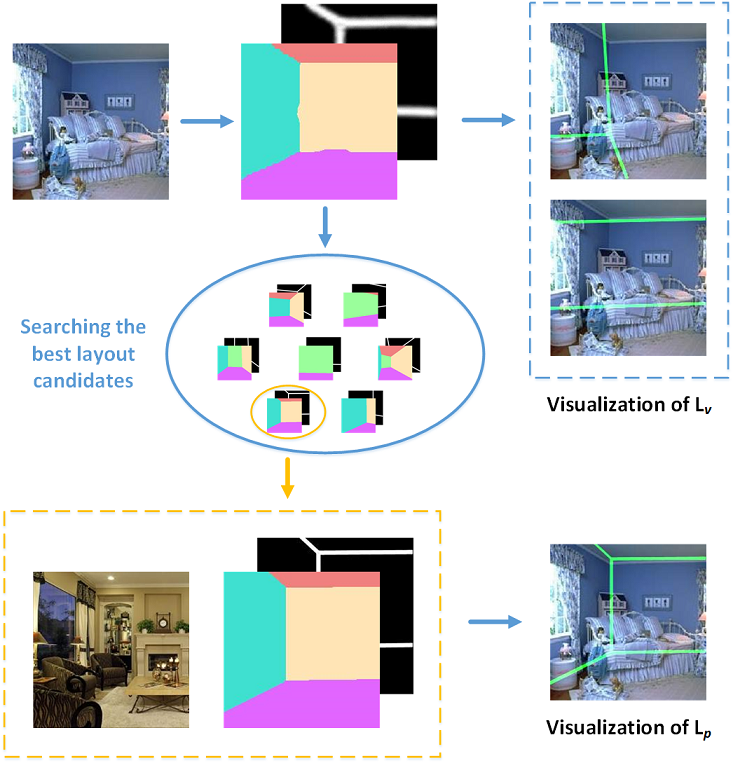}
\end{center}
\caption{An example of a cluttered room. The layouts from sampling are inaccurate, which is caused by the error of the estimated vanishing points. On the contrary, a good layout hypothesis is obtained by searching in the predefined pool.}
\label{fig:suplayout}
\end{figure}

\subsection{Generating layouts from ray sampling}

Based on the ``Manhattan world assumption''~\cite{coughlan2000manhattan}, there exist three orthogonal vanishing points in an indoor scene. A layout can be generated by sampling at most two rays from the vertical and farther horizontal vanishing points. We first estimated the three vanishing points of each image using the method of~\cite{hedau2009recovering}. Then, we ordered the three vanishing points as vertical, farther horizontal, and closer horizontal points.

Mallya and Lazebnik~\cite{mallya2015learning} proposed an adaptive ray-sampling method based on the estimated edge maps. First, the edge map is divided into several uniformly spaced sectors by the sampled rays from a vanishing point. Then, a fixed number of sectors that have strong average edge strength are selected, and rays are densely sampled in the selected sectors to construct the layouts. We modified this method to make the number of selected sectors adaptive, which significantly reduces the occurrence of bad layout hypotheses.

The total number of sectors is denoted as $H$. The average edge strength of each sector is denoted as $d_i, i = 1,\cdots ,H$. In our method, the $i_{th}$ sector is selected only if $i$ satisfies the following two conditions simultaneously:

(i)	$d_i > d_{i+1}$ and $d_i > d_{i-1}$,

(ii) $d_i - d_{i+1} > D$ or $d_i - d_{i-1} > D$.

Considering the sectors on the image boundary, we define $d_0 = 0$ and $d_{H+1} = 0$. The threshold $D$ is set to 0.03 in the experiments. The first condition is used to select the sectors with local maximum edge strength, and the second condition can avoid unnecessary selections caused by noise.


The whole process of generating layouts from ray sampling is shown in Fig.~\ref{fig:samp}. Uniformly spaced sectors are first generated from $vp_1$, and $N$ (e.g., $N$ = 3) rays are uniformly sampled in each selected sector. The similar procedure is performed on $vp_2$. Finally, we generate rays from $vp_3$ to go through the intersection points of the previously sampled rays of $vp_1$ and $vp_2$ to yield a layout candidate. In order to improve the error tolerance, we also generate different types of layouts by using a subset of the selected sectors. In the implementation, we can obtain 73 layouts for each room image on average.

Given the large number of generated layouts, we need to select the best of each type for the subsequent refinement. We used the scoring function Eq. (\ref{eq:score}) to rank the layouts. We selected the $K_{1}$ (e.g., $K_{1}$ = 2) best layouts of different types to form the layout hypotheses $\textbf{L}_{v}$.

\begin{figure}
\begin{center}
\includegraphics[width = .5\columnwidth]{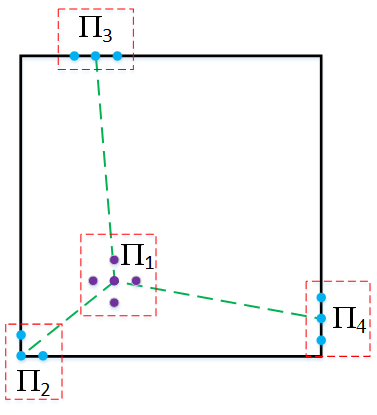}
\end{center}
\caption{Illustration of the neighboring point sets of internal corner point $p_{1}$ and the border corner points $p_{2}$, $p_{3}$ and $p_{4}$.}
\label{fig:refine}
\end{figure}

\begin{algorithm}
\caption{Layout Optimization.} 
\hspace*{0.02in} {\bf Input:} 
layout hypotheses $\textbf{L}$, segmentation map $\textbf{M}$, edge map $\textbf{E}$\\
\hspace*{0.02in} {\bf Output:} 
optimized layout $l^{*}$
\begin{algorithmic}
\For{\textbf{each} $l_{k} \in \textbf{L}, k = 1,2,...,K$} 
　　\State $l = l_{k},\quad \textrm{where} \ l = (t,p_{1},p_{2},...,p_{n})$
    \State $s =S(\textbf{M}_{l},\textbf{E}_{l}|\textbf{M},\textbf{E})$ 
        \While {$True$}
        \For{\textbf{each} $p_{i}$}
            \State generate neighbor points $\Pi_{i} = \{p^{1}_{i},p^{2}_{i},...\}$
                \For{\textbf{each} $p^{j}_{i} \in \Pi_{i}$}
                    \State replace $p_{i}$ with $p^{j}_{i}$, obtain a new layout $l'$
                    \State $s' = S(\textbf{M}_{l'},\textbf{E}_{l'}|\textbf{M},\textbf{E})$
                    \If{$s' > s$}
　　　　                \State $l = l'$,\quad $s = s'$
　　                \EndIf
                \EndFor
        \EndFor
        \If {$\textit{score does not increase}$}
            \State \textbf{break}
        \EndIf
        \EndWhile
    \State $l^{*}_{k} = l$
\EndFor
\State \Return $l^{*}$ = max$(l^{*}_{k})$
\end{algorithmic}
\label{alg:ref}
\end{algorithm}

\subsection{Generating layouts from a predefined pool}

With the exception of the predicted edge maps, the quality of the sampled layouts is directly affected by the estimated vanishing points. Unfortunately, the estimation of the vanishing points will inevitably have an error. A small error can be corrected in the refinement stage, but if the estimated vanishing points deviate far from the truth, the layout hypotheses obtained in $\textbf{L}_{v}$ will be completely inaccurate, and cannot be corrected by refinement. For example, in Fig.~\ref{fig:suplayout}, the estimated vanishing points of a cluttered room are poor, and thus the layouts of $\textbf{L}_{v}$ are inaccurate, even though the network estimates are basically accurate.

With the assumption that the indoor scenes may share similar spatial organization, we propose to introduce an extra set of layout hypotheses to improve the error tolerance, which is formed by searching for layouts in a predefined layout pool. To generate a rich pool, the training samples of the LSUN dataset (scaled to match the square size) are all included, and cover 4000 typical layouts of different indoor scenes.

Each layout sample of the predefined pool is evaluated using the scoring function Eq. (\ref{eq:score}). Only the best layout (with the highest score) of each type is retained, and the top $K_{2}$ (e.g., $K_{2}$ = 2) layouts among these best layouts of different types are selected to form hypotheses $\textbf{L}_{p}$. Because the layout hypotheses will be refined later, it is unnecessary to generate a precise one. If using the proposed hypotheses $\textbf{L}_{p}$ only, the pre-processing steps and data driven proposals are unnecessary. In Fig.~\ref{fig:suplayout}, the layout pool offers a good layout that is very close to the network prediction.


\subsection{Layout optimization}

The final layout hypotheses $\textbf{L}$ are produced by combining the hypotheses $\textbf{L}_{v}$ and hypotheses $\textbf{L}_{p}$, and they thus contain $K = K_{1} + K_{2}$ different layouts.

\begin{equation}
\begin{aligned}
\label{eq:combine}
\textbf{L} = \textbf{L}_{v}\cup \textbf{L}_{p}.
\end{aligned}
\end{equation}

\begin{figure}
\begin{center}
\includegraphics[width = 0.9\columnwidth]{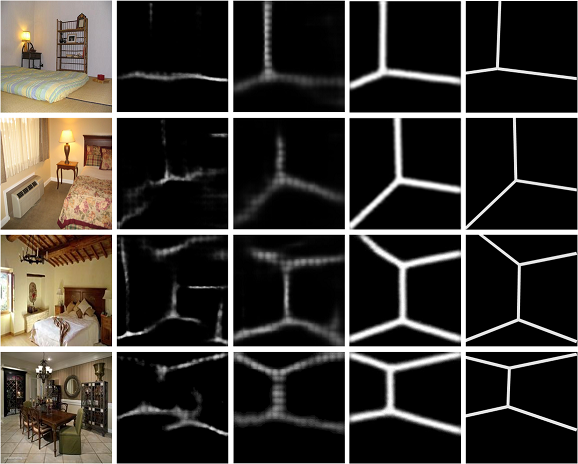}
\end{center}
\caption{Comparison of the estimated edge maps (from left to right): the input image, the estimated edge map obtained by FCN~\cite{mallya2015learning}, MFCN~\cite{ren2016coarse}, the proposed network, and the ground truth.
}
\label{fig:fcncmp}
\end{figure}

\begin{table}[]
\centering
\caption{Evaluation of edge estimation on the Hedau dataset.}
\label{table:ods}
\begin{tabular}{|c|c|c|c|}
\hline
Metrics & FCN~\cite{mallya2015learning}   & MFCN~\cite{ren2016coarse}  & Ours  \\ \hline
ODS     & 0.255 & 0.265 & 0.329 \\ \hline
OIS     & 0.263 & 0.291 & 0.352 \\ \hline
\end{tabular}
\end{table}

\begin{figure*}
\begin{center}
\includegraphics[width = 1.0\textwidth]{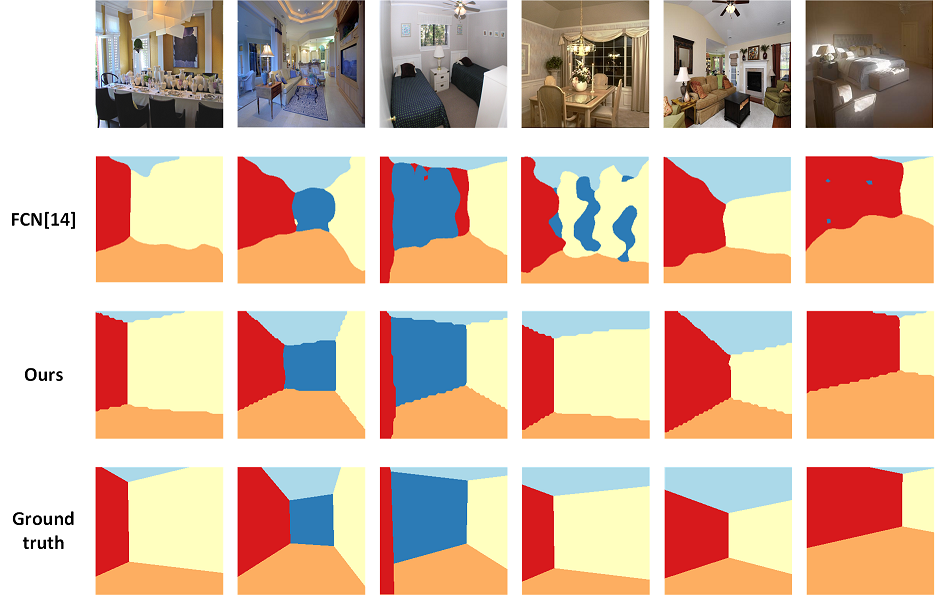}
\end{center}
\caption{Comparison of segmentation maps generated by FCN~\cite{dasgupta2016delay} and the proposed network.
}
\label{fig:seg_comparison}
\end{figure*}

Next, we introduced a refinement step to further optimize the layout hypotheses via a greedy strategy. Given an initial layout, we first rounded the coordinates of the corner points to the nearest integer. Then, we sequentially searched among the neighboring pixels of each corner point and used the one with the highest matching score defined in Eq. (\ref{eq:score}) to update the current corner point. The process is repeated until the score does no longer increase.

For $l \in \textbf{L}$, let $p_{i}$ be a corner point of $l$. We define $\Pi_{i}$ as a point set consisting of the neighboring pixels of $p_{i}$. Specifically, if $p_{i}$ is a corner point on the image boundary, then $\Pi_{i}$ contains $p_{i}$ and its two nearest pixels on both sides of $p_{i}$; if $p_{i}$ is a corner point inside the image, $\Pi_{i}$ is composed of $p_{i}$ and its four nearest pixels (up, down, left, right). An example is given in Fig.~\ref{fig:refine}, where $\Pi_{1}$ is the neighboring point set for an internal corner point $p_{1}$, while $\Pi_{2}$, $\Pi_{3}$, and $\Pi_{4}$ are the neighboring point sets of border corner points $p_{2}$, $p_{3}$, and $p_{4}$, respectively.

Algorithm~\ref{alg:ref} describes the procedure of the proposed layout optimization. Given an initial layout $l$, we first compute the current matching score $s$ using Eq. (\ref{eq:score}). Then, for each corner point $p_{i}$, we replaced $p_{i}$ with a pixel in $\Pi_{i}$ to obtain a new layout $l'$. The score of $l'$ is computed and denoted as $s'$. If $s'$ is higher than $s$, we update $l$ and $s$ with $l'$ and $s'$, respectively. The process is repeated until the score does not change any more. In this way, each layout hypothesis in $\textbf{L}$ is refined individually, and the final optimized layout $l^{*}$ is the one with the highest consistency score.

\begin{table}
\caption{Layout estimation performance on the Hedau dataset.}
\begin{center}
\begin{tabular}{|c|c|}
\hline
Methods&Pixel error (\%) \\ \hline
Hedau \emph{et~al.} (2009)~\cite{hedau2009recovering}&21.20 \\ \hline
Gupta \emph{et~al.} (2010)~\cite{gupta2010estimating}&16.20 \\ \hline
Zhao \emph{et~al.} (2013)~\cite{zhao2013scene}&14.50 \\ \hline
Ramalingam \emph{et~al.} (2013)~\cite{ramalingam2013manhattan}&13.34 \\ \hline
Mallya \emph{et~al.} (2015)~\cite{mallya2015learning}&12.83 \\ \hline
Schwing \emph{et~al.} (2012)~\cite{schwing2012efficient}&12.80 \\ \hline
Del Pero \emph{et~al.} (2013)~\cite{del2013understanding}&12.70 \\ \hline
Dasgupta \emph{et~al.} (2016)~\cite{dasgupta2016delay}&9.73 \\ \hline
Ren \emph{et~al.} (2016)~\cite{ren2016coarse}&8.67 \\ \hline
Lee \emph{et~al.} (2017)~\cite{lee2017roomnet}&8.36 \\ \hline
Ours(trained on Hedau)&7.94 \\ \hline
\textbf{Ours (trained on LSUN)}&\textbf{7.36} \\ \hline
\end{tabular}
\end{center}
\label{table:hedau}
\end{table}

\begin{table}
\caption{Layout estimation performance on the LSUN dataset.}
\begin{center}
\begin{tabular}{|c|c|c|}
\hline
Methods&Pixel error (\%) &Corner error (\%) \\ \hline
Hedau \emph{et~al.} (2009)~\cite{hedau2009recovering} &24.23&15.48 \\ \hline
Mallya \emph{et~al.} (2015)~\cite{mallya2015learning} &16.71&11.02 \\ \hline
Dasgupta \emph{et~al.} (2016)~\cite{dasgupta2016delay}&10.63&8.20 \\ \hline
Ren \emph{et~al.} (2016)~\cite{ren2016coarse}&9.31&7.95 \\ \hline
Lee \emph{et~al.} (2017)~\cite{lee2017roomnet}&9.86&6.30 \\ \hline
\textbf{Ours}&\textbf{6.58}&\textbf{5.17} \\ \hline
\end{tabular}
\end{center}
\label{table:lsun}
\end{table}

%
%
%

\begin{table}[]
\centering
\captionsetup{justification=raggedright}
\caption{Evaluation of joint training and separate training on the LSUN validation dataset. }
\label{table:train}
\begin{tabular}{|c|c|c|c|c|}
\hline
\multirow{2}{*}{\begin{tabular}[c]{@{}c@{}} \\  Training methods \end{tabular}} & \multicolumn{2}{c|}{\begin{tabular}[c]{@{}c@{}}Evaluation of the \\ network output\end{tabular}}                                                            & \multicolumn{2}{c|}{\begin{tabular}[c]{@{}c@{}}Layout estimation \\ performance\end{tabular}}                          \\ \cline{2-5}
                                  & \begin{tabular}[c]{@{}c@{}}Edge error\end{tabular} & \begin{tabular}[c]{@{}c@{}}Semantic \\ error (\%)\end{tabular} & \begin{tabular}[c]{@{}c@{}}Pixel \\ error (\%)\end{tabular} & \begin{tabular}[c]{@{}c@{}}Corner \\ error (\%)\end{tabular} \\ \hline
Edge maps only                    & 11.36                                                                       & -                                                                             & 10.45                                                     & 7.73                                                       \\ \hline
Semantic labels only              & -                                                                           & 9.72                                                                          & 9.14                                                      & 7.12                                                       \\ \hline
Joint learning                    & 11.32                                                                       & 7.91                                                                          & 6.94                                                      & 5.16                                                       \\ \hline
\end{tabular}
\end{table}

\begin{table*}[]
\centering
\caption{Evaluation of different layout generation methods on the Hedau dataset and the LSUN validation dataset (\%).}
\label{table:lals}
\begin{tabular}{|c|c|c|c|c|c|c|c|c|}
\hline
\multirow{3}{*}{Set of layouts}    & \multicolumn{2}{c|}{Hedau dataset}   & \multicolumn{6}{c|}{LSUN validation dataset}                                                                          \\ \cline{2-9}
                                   & Before refinement & After refinement & \multicolumn{3}{c|}{Before refinement}                    & \multicolumn{3}{c|}{After refinement}                     \\ \cline{2-9}
                                   & Pixel error   & Pixel error  & Pixel error  & Corner error  & Type accuracy  & Pixel error  & Corner error  & Type accuracy  \\ \hline
$\textbf{L}_{v}$                   & 9.44              & 8.40             & 8.80             & 6.52              & 74.11              & 7.57             & 5.98              & 74.87              \\ \hline
$\textbf{L}_{p}$                   & 8.95              & 7.40             & 8.90             & 6.78              & 72.59              & 7.10             & 5.91              & 75.63              \\ \hline
$\textbf{L}_{v}\cup\textbf{L}_{p}$ & \textbf{8.18}   & \textbf{7.36}  & \textbf{8.02}  & \textbf{5.84}   & \textbf{76.90}              & \textbf{6.94}  & \textbf{5.16}   & \textbf{81.22}              \\ \hline
\end{tabular}
\end{table*}

\begin{figure}
\begin{center}
\includegraphics[width = 0.95\columnwidth]{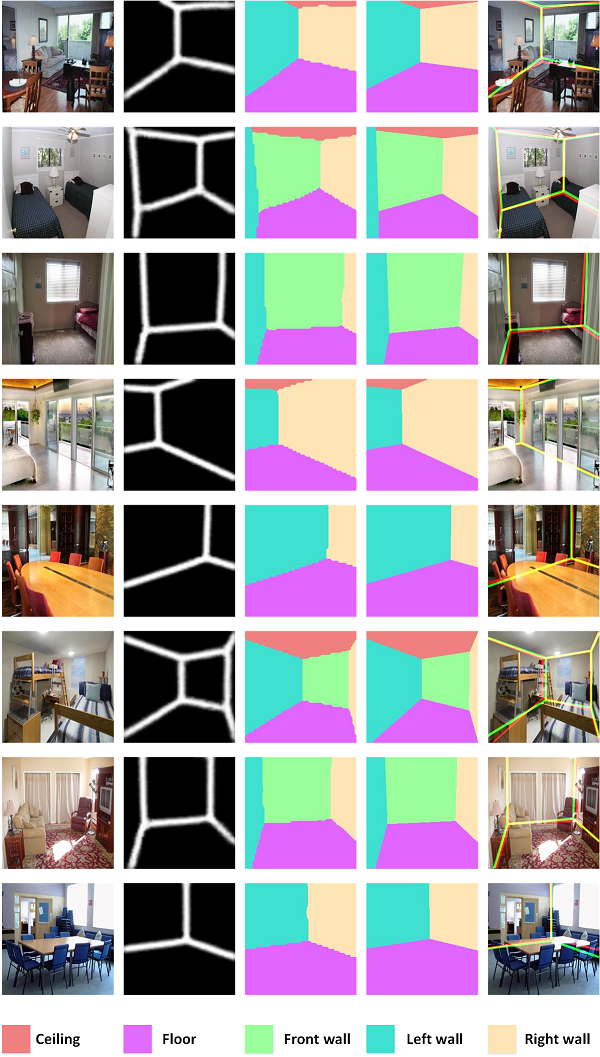}
\end{center}
\caption{Layout estimation results on the LSUN validation set (from left to right): the input image, estimated edge map, estimated segmentation map, ground-truth segmentation map, and a comparison of the layout output (outlined by green) and the ground truth (outlined by red).}
\label{fig:good}
\end{figure}

\section{Experimental results}

We performed the experiments on two benchmark layout datasets: Hedau~\cite{hedau2009recovering} and LSUN~\cite{yinda2016lsun}. We obtained the Hedau dataset from the Web and from LabelMe~\cite{russell2008labelme}, which consists of 209 training images and 105 testing images. The LSUN was developed recently for scene-centric large-scale challenges. It is much larger than Hedau, and contains 4000 training images, 394 validation images, and 1000 testing images. All of the images have valid room layouts that can be clearly annotated manually. There are eight scene categories in LSUN, including a bedroom, hotel room, dining room, dinette home, living room, office, conference room, and classroom. Similar to ImageNet~\cite{deng2009imagenet}, the layout annotations of the test images are not provided, and thus the results should be sent to the organizers for evaluation.

While training the deconvolution network, we augmented the 4000 training samples of LSUN by cropping, horizontal flipping, and color jittering. Because the number of training samples of Hedau was too small, we did not train using this dataset, and performed testing using only the model trained on LSUN.

The proposed algorithm was implemented with Matlab R2015a on a PC with Intel i5-4590 CPU (3.30-GHz). The estimation of edge map and semantic labels takes 0.07s, and the layout hypotheses $\textbf{L}_{v}$ and $\textbf{L}_{p}$ take 37.28s and 34.90s, respectively. The refinement step is slower and takes 77.93s, as it runs in pixel level.

\subsection{Layout estimation performance}

\begin{figure}
\begin{center}
\subfigure[]{
\includegraphics[width = 0.95\columnwidth]{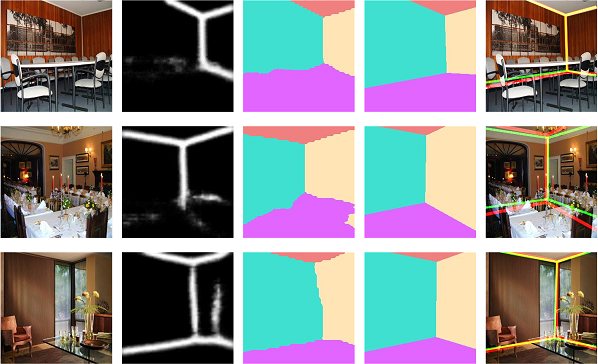}
}
\subfigure[]{
\includegraphics[width = 0.95\columnwidth]{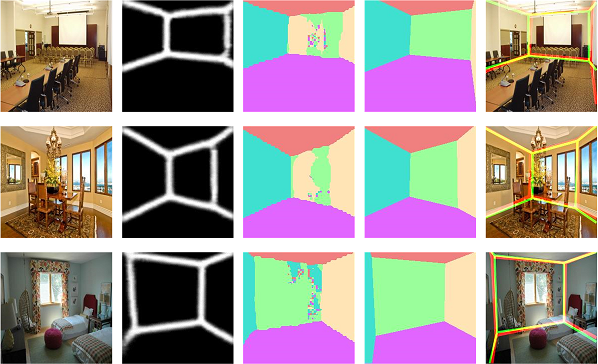}
}
\subfigure[]{
\includegraphics[width = 0.95\columnwidth]{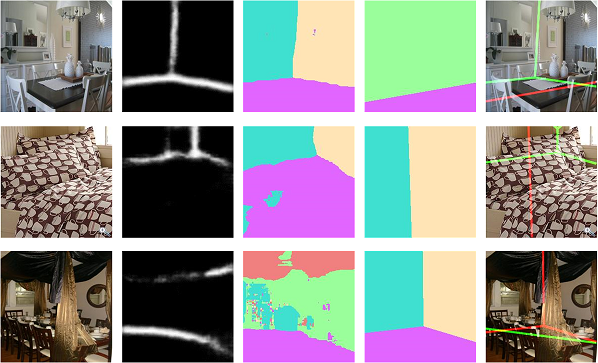}
}
\end{center}
\caption{Some examples from the LSUN validation set with poor network outputs.}
\label{fig:bad}
\end{figure}

In Fig.~\ref{fig:fcncmp}, we compared the edge maps generated from FCN~\cite{mallya2015learning}, MFCN~\cite{ren2016coarse}, and the proposed deconvolution network. It can be concluded that the edge maps obtained from the proposed network are clearer and more precise than those obtained from FCNs. The quantitative comparison of the edge map estimation is given in Table~\ref{table:ods}. As in~\cite{mallya2015learning} and~\cite{ren2016coarse}, the evaluation metrics are the fixed contour threshold (ODS) and the per-image best threshold (OIS)~\cite{arbelaez2011contour}. It is observed that the proposed network yielded the best ODS and OIS scores, which indicates its superiority over FCN and MFCN in edge prediction. The comparison of the estimated segmentation maps are shown in Fig.~\ref{fig:seg_comparison}. Apparently, the proposed network produced more precise segmentation maps than those obtained from FCN~\cite{dasgupta2016delay}. There are two reasons: First, the proposed network has fully connected layers for larger receptive field, and thus can better take advantage of the global context; Second, the semantic labels are jointly learned with the edge maps, where they could mutually benefit each other.

Some layout estimation results are shown in Fig.~\ref{fig:good}. As can be seen, the final layout estimates are very close to the ground truth. The figure shows that the estimated edge maps and segmentation maps are very robust to occlusion and clutter. As an example, consider the room in the first, where the boundary between the left wall and floor is completely occluded by the sofa and the table. Moreover, there is an open door on the left wall which may generate strong edges. However, our network still produced accurate layout estimates in this challenging scene.

Because the edge map and segmentation map are estimated individually with two subnetworks, poor estimates will be obtained on either of them in some cases, as shown in Fig.~\ref{fig:bad} (a) and (b). In (a), some edges are missed in the first two examples caused by severe occlusion of cluttered rooms. In the third row, some layout edges are wrongly detected owing to the influence of the strong structure of a wall in the scene. In (b), the segmentation maps are poor and have multiple disjoint labels for the same wall. Such errors may be incurred by the labeling ambiguity of walls. Although either the edge map or segmentation map is poor in the above two cases, our proposed method still produced accurate layout estimates. This proves that the edge estimation and segmentation estimation could compensate each other, which makes the algorithm robust to various scenarios. Fig.~\ref{fig:bad} (c) gives the worst cases when the estimations of both edge maps and segmentation maps are poor.

We also compared our method with state-of-the-art layout estimation work, as shown in Table~\ref{table:hedau} and Table~\ref{table:lsun}. We used the pixel error and corner error for the quantitative evaluation. The pixel error reflects the misclassification rate of the pixels, while the corner error reflects the positional error of the corner points between the layout and ground truth. Our method achieves the best performance for both datasets, and outperforms the second best result by 2.73\% of the pixel error and 1.13\% of the corner error on the LSUN dataset, and 1.00\% of the pixel error on the Hedau dataset.


%


Besides the studies in Fig.~\ref{fig:bad}, we intend to further investigate the effects of joint learning the room edges and semantic labels compared to learning each of them separately. For this purpose, we trained two extra networks on the LSUN dataset: The first one is trained with edge maps only, and the second one is trained with the semantic labels only. As shown in Table~\ref{table:train}, evaluation was performed on the LSUN validation dataset to compare the network outputs and the layout error. The edge error is defined by the Euclidean distance between the estimated edge map and ground truth, while the semantic error is defined by the bipartite matching with the minimal pixel-wise error between the estimated segmentation map and ground truth. The results are shown in Table~\ref{table:train}. Apparently, joint training produced the lowest edge error, semantic error and layout error, which proved the superiority of joint training.



Finally, we compared the two layout hypotheses in Table~\ref{table:lals}. The first two rows show the performance obtained when using the layout hypotheses of $\textbf{L}_{v}$ and $\textbf{L}_{p}$. Also, the layout type accuracy is introduced as an additional metric for evaluation on the LSUN validation dataset. Apparently, both the layout error and the type prediction accuracy improve significantly when $\textbf{L}_{p}$ is added to $\textbf{L}_{v}$. The reason might be that humans have similar viewpoint preferences when shooting indoor scenes, and thus the room layouts share many similarities. The evidence is that the layouts of about $70\%$ of room images in the LSUN dataset belong to type 4 or type 5 (see Fig.~\ref{fig:type}). Such a similarity is further promoted as the layouts are all resized to a fixed scale in the algorithm. However, if an indoor scene image is shot from an unusual viewpoint, a proper layout cannot be given from the pool. It is necessary to generate layout hypotheses by ray sampling, especially for robotics applications for which the images may be captured from any viewpoint. In addition, Table~\ref{table:lals} also demonstrates the accuracy gains realized by the proposed layout optimization strategy.

\section{Conclusions}

In this paper, we presented a framework to estimate the spatial layout from a single image. We trained a network with shared encoder and two separate decoders to jointly predict the edge maps and semantic labels. The two outputs are jointly learned not only to enhance the training, but also to compensate each other for layout estimation. In addition, we proposed a more elegant layout generation and refinement strategy to make the layout estimation more accurate and robust. Extensive experimental results obtained using benchmark datasets demonstrated the superiority of the proposed method over existing layout estimation works.



%

%
%
%
%
%

\ifCLASSOPTIONcaptionsoff
  \newpage
\fi



%
%
%

\bibliographystyle{IEEEtran}

\end{document}